\documentclass{article}

\usepackage{microtype}
\usepackage{graphicx}
\usepackage{subcaption}
\usepackage{booktabs}
\usepackage{multirow}
\usepackage{hyperref}
\usepackage{xurl}

\usepackage[preprint]{icml2026}

\usepackage{amsmath}
\usepackage{amssymb}
\usepackage{mathtools}
\usepackage[capitalize,noabbrev]{cleveref}
\usepackage{xcolor}
\usepackage{xspace}
\usepackage{listings}
\usepackage{enumitem}
\usepackage{float}
\usepackage{needspace}

\setlist[itemize]{leftmargin=1.2em, topsep=3pt, itemsep=2pt, parsep=0pt, partopsep=0pt}

\newcommand{\ours}{Ours\xspace}
\newcommand{\oursL}{Ours (LIBERO only)\xspace}
\newcommand{\rul}{RURL\xspace}
\newcommand{\pizf}{$\pi_{0.5}$\xspace}
\newcommand{\task}[1]{\texttt{\small #1}}
\makeatletter
\newcommand{\icmlaffiliationorder}[1]{\setbox0\hbox{\@for\icml@aff:=#1\do{\@pa{\icml@aff}}}}
\makeatother

\definecolor{codebg}{HTML}{F5F5F2}
\lstdefinestyle{prompt}{basicstyle=\fontencoding{T1}\fontfamily{lmtt}\footnotesize\selectfont, columns=fullflexible, keepspaces=true, upquote=true, breaklines=true,
        breakatwhitespace=true, breakindent=0pt, backgroundcolor=\color{codebg}, frame=single, framerule=0pt, framesep=5pt,
        rulecolor=\color{codebg}, xleftmargin=5pt, xrightmargin=5pt, aboveskip=6pt, belowskip=6pt}

\icmltitlerunning{Training Reasoning Models in Simulation to Natively Use Real Robots}

\begin{document}

\twocolumn[
  \icmltitle{Training Reasoning Models in Simulation to Natively Use Real Robots}

  \icmlsetsymbol{equal}{*}
  \icmlsetsymbol{incowork}{\textdagger}
  \icmlsetsymbol{nvidiawork}{\textdaggerdbl}
  \icmlaffiliationorder{mit,berkeley,ucsd,nvidia,inco}
  \begin{icmlauthorlist}
    \icmlauthor{Jiaming Tang}{equal,mit,inco,incowork}
    \icmlauthor{Zhuoyang Zhang}{equal,mit}
    \icmlauthor{Junxian Guo}{equal,mit,nvidiawork}
    \icmlauthor{Fangzhou Zhao}{berkeley,inco}
    \icmlauthor{Yilong Zhao}{berkeley}
    \icmlauthor{Zekai Li}{ucsd}
    \icmlauthor{Xingyang Li}{mit}
    \icmlauthor{Yuyang Zhao}{nvidia}
    \icmlauthor{Song Bian}{nvidia}
    \icmlauthor{Zhijian Liu}{ucsd,inco}
    \icmlauthor{Song Han}{mit,nvidia}
    \\
    \rule{0pt}{11pt}\url{https://github.com/mit-han-lab/pear}
  \end{icmlauthorlist}
  \icmlaffiliation{mit}{MIT}
  \icmlaffiliation{berkeley}{UC Berkeley}
  \icmlaffiliation{ucsd}{UCSD}
  \icmlaffiliation{nvidia}{NVIDIA}
  \icmlaffiliation{inco}{Inco AI}
  \icmlcorrespondingauthor{Song Han}{songhan@mit.edu}

  \icmlkeywords{robot learning, reinforcement learning, vision-language models, sim-to-real, robot-use agents}

  \vskip 0.2in
  \begin{center}
  \includegraphics[width=\textwidth]{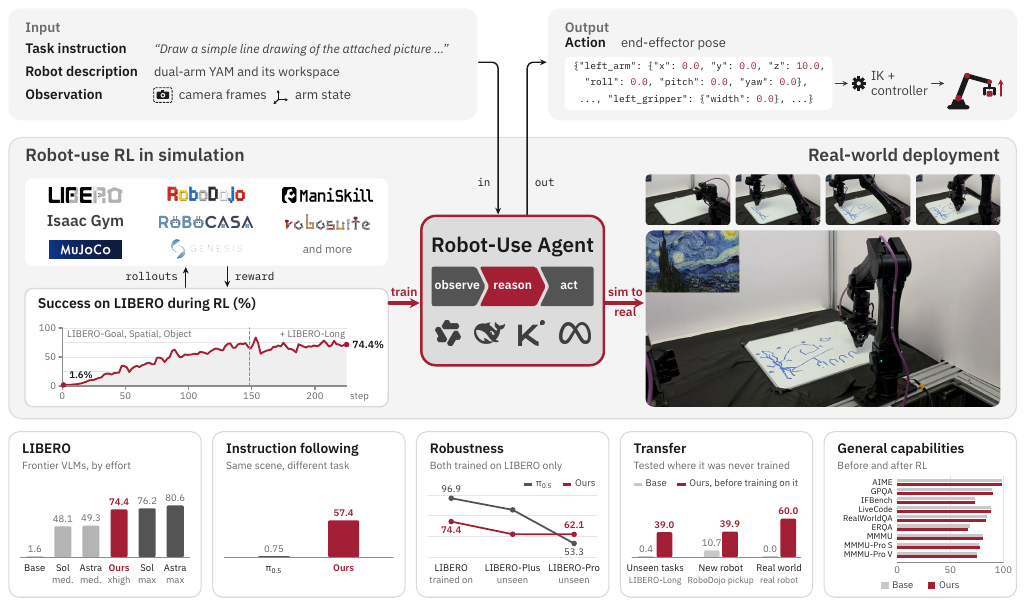}
  \captionof{figure}{\textbf{Overview of robot-use reinforcement learning (RURL).}
  Like a computer-use agent, the robot-use agent (RUA) observes the scene, reasons in a chain of thought, and generates the next end-effector pose (top).
  It is trained with reinforcement learning entirely in simulation (left) and deployed on a real robot (right).
  The logos show examples of simulators and open-weight models; we train Qwen3.8-27B in LIBERO and RoboDojo.
  Bottom: results on LIBERO, instruction following, robustness, transfer to unseen tasks, a new robot and the real world, and general capabilities, all without a harness.}
  \label{fig:method}
\end{center}

  \vskip 0.1in
]

\printAffiliationsAndNotice{\icmlEqualContribution \textsuperscript{\textdagger}Part of this work was done during an internship at Inco AI. \textsuperscript{\textdaggerdbl}Part of this work was done during an internship at NVIDIA. }

\begin{abstract}
Robot-use agents (RUAs) have emerged as a promising approach to general-purpose robot control.
However, strong performance has so far been achieved only by frontier proprietary models such as GPT-6 Astra and Claude Fable 5.2, whereas open-weight models remain substantially less capable.
We show that an open-weight 27B reasoning model can be trained to use real robots natively through reinforcement learning, entirely in simulation and without demonstrations.
Like a computer-use agent (CUA), the model observes the environment, reasons in a chain of thought, and generates the next end-effector pose, without any harness.
We build a Post-training Engine for Agentic Robotics (PEAR) and use it to train the model on simulated robot tasks at scale with robot-use reinforcement learning (RURL), mirroring how frontier models acquire computer use.
We find that the skills acquired in simulation generalize well beyond the training tasks: they transfer to an unseen robot, to the real world, and to novel instructions.
After training on LIBERO and RoboDojo in simulation, the model succeeds in 60\% of our trials on a real dual-arm robot, whereas the base model succeeds in none.
On LIBERO, training raises the success rate from 1.6\% to 74.4\%, approaching GPT-6 Astra and GPT-6.1 Sol at maximum reasoning effort.
Compared with \pizf, a vision-language-action model fine-tuned on demonstrations, our model is substantially more robust to changes in the task itself.
On LIBERO-Pro, \pizf succeeds in only 0.75\% of episodes when assigned a new task in a familiar scene whereas our model succeeds in 57.4\%.
Finally, the model retains its general capabilities, with no measurable degradation on nine reasoning, coding and vision benchmarks.
These results highlight robot-use reinforcement learning as a promising component of foundation-model post-training.
\end{abstract}

\begin{figure*}[t]
  \centering
  \includegraphics[width=\textwidth]{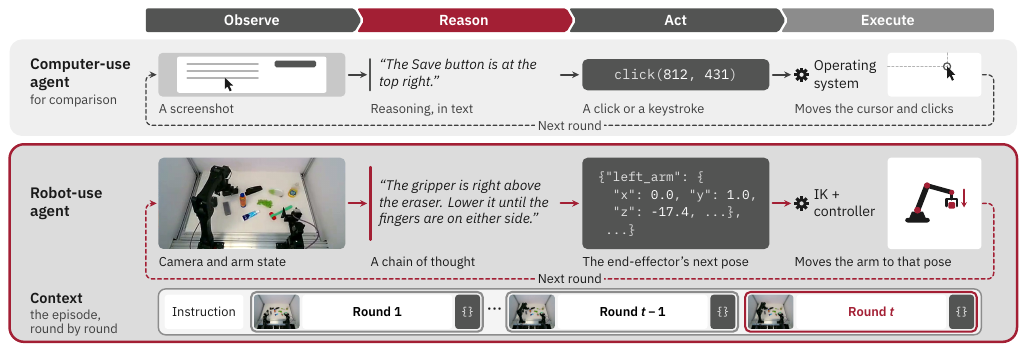}
  \caption{\textbf{One round of robot use.}
  A computer-use agent (top) observes a screenshot, reasons about the task and outputs a click or a keystroke, which the operating system executes.
  Our model (bottom) operates a robot in a similar way: it observes the camera views and the arm's state, reasons in a chain of thought and outputs the end-effector's next pose, and inverse kinematics and the arm's controller then move the arm.
  Each round is added to the context, which also holds the instruction and all earlier rounds.
  The same steps are used in simulation and on the real robot.}
  \label{fig:loop}
\end{figure*}

\begin{figure*}[t]
  \centering
  \includegraphics[width=\textwidth]{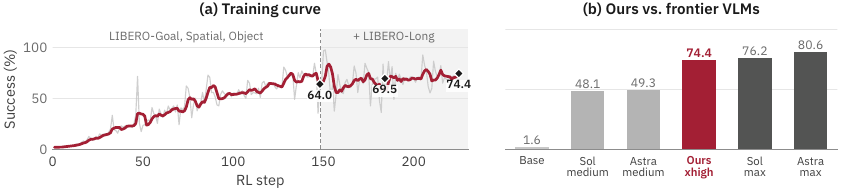}
  \caption{\textbf{Training curve on LIBERO and comparison with frontier VLMs.}
  (a) The success rate of training rollouts per RL step (thin line) and smoothed (thick line); LIBERO-Long is added at step 149, and the diamonds are the same run evaluated on LIBERO's 2,000 episodes.
  (b) The base model, our model at xhigh, the effort used in training, and GPT-6.1 Sol~\citep{gpt61sol} and GPT-6 Astra~\citep{gpt6astra} at medium and max effort, on LIBERO's 2,000 evaluation episodes; all models use the same protocol, and effort is each model's own setting for how long it reasons.}
  \label{fig:libero}
\end{figure*}

\begin{figure*}[t]
  \centering
  \includegraphics[width=\textwidth]{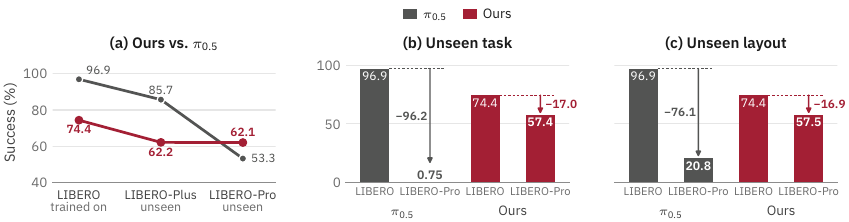}
  \caption{\textbf{Comparison with \pizf on LIBERO, LIBERO-Plus and LIBERO-Pro.}
  Both policies are trained only on LIBERO: \pizf with supervised fine-tuning on demonstrations and our model with RURL.
  (a) Success rate on LIBERO and on the two perturbed benchmarks.
  (b, c) Two of LIBERO-Pro's perturbations, an unseen task in a training scene and swapped object positions; the arrow gives the drop from LIBERO.
  The \pizf results are from prior work~\citep{openpi,flexpi,liberopro}.}
  \label{fig:robust}
\end{figure*}

\begin{figure*}[t]
  \centering
  \includegraphics[width=\textwidth]{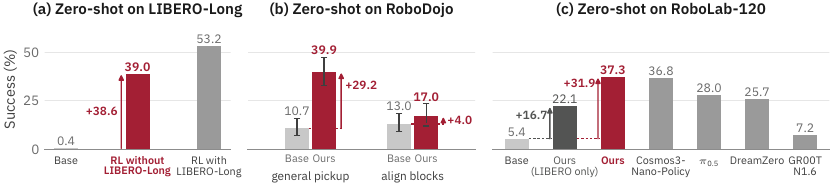}
  \caption{\textbf{Zero-shot success rate on unseen tasks, a new simulator and robot, and a new benchmark.}
  (a) RL without LIBERO-Long is trained on the other three LIBERO suites; RL with LIBERO-Long, the same run after LIBERO-Long is added, is shown for reference.
  (b) RoboDojo, a different simulator with a dual-arm ARX, after RL on LIBERO only; error bars are 95\% Wilson intervals.
  (c) RoboLab-120, a third benchmark, against four generalist robot policies from the RoboLab leaderboard~\citep{robolableaderboard}, all DROID~\citep{droid} checkpoints.
  }
  \label{fig:transfer}
\end{figure*}

\section{Introduction}
\label{sec:intro}

Robot-use agents (RUAs) have recently become a promising direction for general-purpose robot control~\citep{isola,ora,rpent,aspire}.
Similar to a computer-use agent (CUA) that operates a computer from screenshots~\citep{cua,claudecu}, an RUA uses a vision-language model (VLM) that observes the robot's camera images, reasons about the task, and outputs the next end-effector pose.
Because the VLM operates the robot directly, RUAs take advantage of the broad world knowledge and the strong reasoning ability of a capable foundation model, which allows them to generalize and to complete complex tasks.

RUAs built on frontier models such as GPT-6 Astra~\citep{gpt6astra,zap,roboicl} and Claude Fable 5.2~\citep{fable52} have shown exceptional generalization and language following.
On RoboDojo~\citep{robodojo,astrarobodojo} and RoboLab~\citep{robolab,astranomy,sensenova}, agents built on GPT-6 Astra have outperformed the best previous policies.
In contrast, vision-language-action models (VLAs) ~\citep{rt2,openvla,pi0,pi05} fine-tuned on task-specific demonstrations often struggle to generalize to new goals and follow novel instructions~\citep{rl4vla,liberopro,liberoplus,liberocf,liberopara,langgap,igar}. Two factors may contribute to the performance of frontier RUAs: the models have been trained for computer use, and they have strong spatial reasoning abilities.
We hypothesize that the ability to reason in a chain of thought is particularly important for generalization.

However, such performance has so far been obtained only with the strongest proprietary models.
Open-weight models have low success rates on the same benchmarks and fail on some simple tasks as well.
Several works build a harness that adds skills and components such as the segmentation model SAM~3~\citep{sam3} and turns some motions into a small set of options~\citep{ora,aspire,rpent,openrua,rax,pyrualean,robodojoagent}.
Such a harness allows open-weight models to complete some of the tasks~\citep{ora}, but it may restrict their action space and thereby make other tasks infeasible.
As models become more capable, some components of the harness may also be absorbed into the models themselves.

Computer-use agents such as OpenAI's are trained with reinforcement learning (RL) on computer tasks~\citep{cua,operatorcard,uitars2}, and the close resemblance between the two settings suggests that robot use could be learned in the same way.
We therefore train an open-weight reasoning model with robot-use reinforcement learning (RURL), entirely in simulation and without demonstrations. To support this long-horizon, multi-turn \rul at scale, we build PEAR (a Post-training Engine for Agentic Robotics), an infrastructure designed specifically for training RUAs (\cref{sec:pear}). In each round, the model observes camera images and the robot’s state, reasons about the task, and outputs the next end-effector pose (\cref{fig:method}). Inverse kinematics converts this pose into joint targets, and the robot’s controller executes the movement before the model receives the next observation. Earlier rounds remain in the context, and each episode receives a binary reward indicating whether the task was completed (\cref{fig:loop}). We provide no perception tools, planners, or skill libraries, so the model itself must localize objects, determine how to manipulate them, and recognize when an action has failed. Because the model specifies poses rather than joint commands and each robot's own inverse kinematics and controller execute them, the same interface applies in simulation and on a real robot.

Although the model is trained entirely in simulation, the skills it learns transfer well to the real world.
We train the model first on LIBERO~\citep{libero} and then on two tasks in RoboDojo~\citep{robodojo}.
On a real dual-arm robot that it never encountered during training, the model then completes 60\% of our trials, whereas the base model completes none (\cref{sec:sim2real}).
The checkpoint trained only on the single-arm LIBERO tasks also succeeds in 23\% of the trials. These results suggest that simulation-based RL can equip general-purpose reasoning models with robot-use skills that transfer to unseen real-world embodiments.

\begin{figure*}[t]
  \centering
  \includegraphics[width=\textwidth]{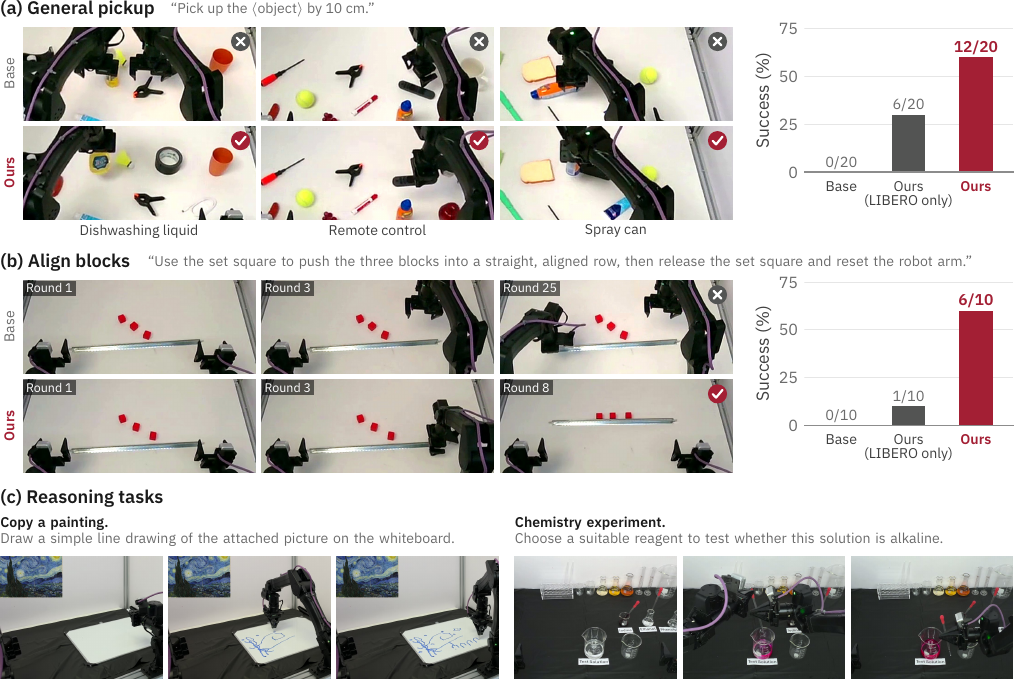}
  \caption{\textbf{Real-robot evaluation on a dual-arm YAM.}
  Our training does not include this robot and uses no real-robot data.
  (a) General pickup: three trials in which the final model picks up the object and the base model does not; the bar chart gives each model's successful trials out of 20, two for each of ten objects.
  (b) Align blocks: three rounds of one trial in which the final model succeeds and the base model fails; the bar chart gives each model's successful trials out of ten.
  (c) Reasoning tasks: the final model draws a simple line version of a painting and selects a reagent that shows whether a solution is alkaline.}
  \label{fig:real}
\end{figure*}

Our main findings are as follows:
\begin{itemize}
  \item \textbf{Approaching frontier models on LIBERO.} The model improves from 1.6\% to 74.4\% on LIBERO, outperforms GPT-6.1 Sol~\citep{gpt61sol} and GPT-6 Astra at medium effort (48.1\% and 49.3\%), and closes most of the gap to them at maximum effort (76.2\% and 80.6\%; \cref{sec:libero}).
  \item \textbf{Robustness and instruction following.} The model remains robust when its training tasks are perturbed and follows instructions it has never seen.
  LIBERO-Pro~\citep{liberopro} changes LIBERO's goals and layouts, and on this benchmark the model outperforms the VLA \pizf~\citep{pi05} (62.1\% vs.\ 53.3\%; \cref{sec:robust}).
  \item \textbf{Zero-shot transfer to unseen tasks and robots.} Without any training on LIBERO-Long, the model reaches 39.0\% on this suite, on which the base model succeeds in 0.4\% of episodes.
  On two RoboDojo tasks with a dual-arm robot, the model trained only on LIBERO succeeds in 39.9\% and 17.0\% of episodes, compared with 10.7\% and 13.0\% for the base model.
  On RoboLab-120, training only on LIBERO raises the success rate from 5.4\% to 22.1\%, and the model trained on LIBERO and RoboDojo outperforms the \pizf entry on the RoboLab leaderboard (37.3\% vs.\ 28.0\%; \cref{sec:transfer}).
  \item \textbf{Sim-to-real transfer.} The skills also transfer zero-shot from simulation to the real world: without any real-robot data, the model trained on LIBERO and RoboDojo succeeds in 60\% of our trials on a real dual-arm robot and the model trained only on LIBERO in 23\%, whereas the base model succeeds in none (\cref{sec:sim2real}).
  Beyond these trials, the model also performs reasoning tasks unlike any seen during training: it draws a simple line version of a painting on a whiteboard and selects a suitable reagent to test whether a solution is alkaline (\cref{fig:real}c).
  \item \textbf{Retained general capabilities.} Robot-use training causes no measurable loss in performance on nine reasoning, coding and vision benchmarks (\cref{sec:general}).
  This result suggests that \rul can be included as a component of foundation-model post-training.
\end{itemize}

\begin{figure*}[t]
  \centering
  \includegraphics[width=\textwidth]{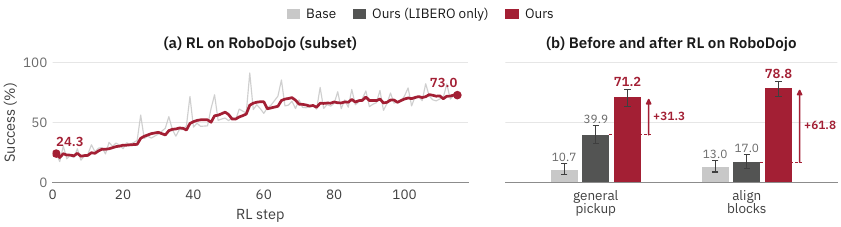}
  \caption{\textbf{Training curve on RoboDojo subset and success rate before and after RL.}
  The final model continues from the LIBERO-only model with RURL on two RoboDojo tasks, entirely in simulation.
  (a) Success rate of training rollouts per RL step (thin line) and smoothed (thick line).
  (b) Success rate on the two tasks before and after this training; error bars are 95\% Wilson intervals.}
  \label{fig:sim2real}
\end{figure*}

\section{Related Work}
\label{sec:related}

\textbf{Robot-use agents.} Frontier VLMs such as GPT-6 Astra~\citep{gpt6astra} have been evaluated as RUAs in simulation and on real robots~\citep{astrarobodojo,astranomy,sensenova,bitterlesson,roboicl}, and RoboDojo~\citep{robodojo} ranks robot agents in a separate track~\citep{robodojoagent}.
Many recent works build RUA harnesses that provide perception tools, skill libraries, motion primitives, memory or code execution, including ORA~\citep{ora}, ASPIRE~\citep{aspire}, Harness VLA~\citep{rpent}, OpenRUA~\citep{openrua}, RAX~\citep{rax}, uni-robot-agent~\citep{unirobotagent} and PyRUA-Lean~\citep{pyrualean}.
Concurrent work also trains open-weight models inside such harnesses, with RL in PyRUA-Lean and with supervised fine-tuning followed by multi-turn RL in CaPtain~\citep{captain}, AgenticRoboRL~\citep{agenticroborl} and ZAP~\citep{zap}.
In contrast, our model uses no harness and no demonstrations.
It is trained with RL alone and transfers from simulation to a real robot.

\textbf{RL for computer-use agents.} Computer-use agents operate a computer from screenshots~\citep{cua,claudecu,uitars} and are evaluated on benchmarks such as OSWorld~\citep{osworld} and WebArena~\citep{webarena}.
OpenAI reports that its agent is trained with supervised learning followed by RL~\citep{operatorcard}.
Open-weight agents have been trained with online multi-turn RL on Android devices in DigiRL~\citep{digirl}, on the web in WebRL~\citep{webrl} and on desktop operating systems in UI-TARS-2~\citep{uitars2}.
RL with verifiable rewards is also widely used to train long chains of thought~\citep{deepseekmath,deepseekr1,dapo}.
RL on a new task erodes a model's other skills less than supervised fine-tuning does~\citep{rlrazor}, and it generalizes better~\citep{sftmemorizes}.
\rul applies this recipe to simulated robot tasks and uses the reasoning model itself as the policy.

\begin{figure}[t]
  \centering
  \includegraphics[width=\linewidth]{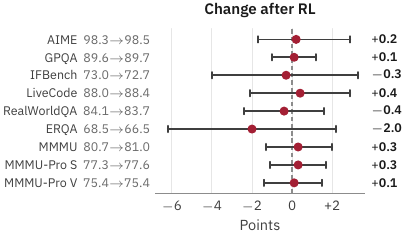}
  \caption{\textbf{General benchmarks before and after RL.}
  The final model against the base model on the same questions: the change in score (dots) with its paired 95\% interval (lines), and the scores before and after RL beside each name.
  The benchmarks are AIME 2026, GPQA Diamond, IFBench, LiveCodeBench v6 (LiveCode), RealWorldQA, ERQA, the MMMU validation set and MMMU-Pro in its standard (S) and vision (V) settings; the mean change over the nine is $-0.15$ points.
  \Cref{tab:general-items} gives the full numbers and the evaluation settings.}
  \label{fig:general}
\end{figure}
\begin{figure}[t]
  \centering
  \includegraphics[width=\linewidth]{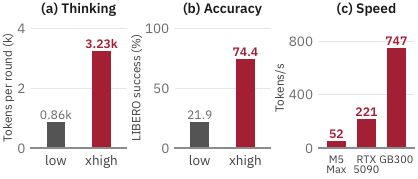}
  \caption{\textbf{Reasoning effort and inference speed.}
  (a, b) Thinking tokens per round and success rate on LIBERO of the LIBERO-only model at low and xhigh effort, with BF16 weights; the low-effort values are the mean of two evaluation runs.
  (c) Decoding speed on one device, with NVFP4 weights quantized with NVIDIA ModelOpt~\citep{modelopt} and Inco's DFlash2 drafter throughout; our megakernel on the GPUs and Inco's Splash engine on the M5 Max.}
  \label{fig:effort}
\end{figure}

\section{Robot-Use Reinforcement Learning}
\label{sec:method}

Our goal is to train a reasoning model in simulation to use a real robot.
We first describe how a robot-use agent operates a robot and the protocol that we use in simulation and on the real robot (\cref{sec:interface}).
We then present our training infrastructure PEAR and the RL algorithm that it runs (\cref{sec:pear}).
Because the base model rarely succeeds at first, training follows a curriculum (\cref{sec:curriculum}).

\subsection{Robot-use agent}
\label{sec:interface}

A computer-use agent~\citep{cua,claudecu} observes a screenshot, reasons about the task, and outputs a click or a keystroke, which the operating system then executes (\cref{fig:loop}, top).
Our model operates a robot~\citep{isola} in the same way (\cref{fig:loop}, bottom).
In each round, it observes the camera frames and the arm's state, reasons in a chain of thought, and outputs the end-effector's next pose.
Inverse kinematics then converts this pose into joint angles, and the arm's controller executes the move.

We use no perception tools, grasp or motion planners~\citep{ora}, or skill libraries~\citep{rpent,aspire}, so the model itself has to localize the objects in the camera frames, determine how to reach and manipulate them, and detect when a move has failed.
Earlier rounds and the instruction remain in the context, so each decision can take the entire episode into account.
The model acquires all of these abilities through RL on robot tasks in simulation (\cref{sec:pear}).

The system prompt states the protocol of this interaction, describes the robot's coordinate frame and cameras, and gives the task instruction (\cref{app:prompts}).
The protocol describes a robot as a set of named parts with named quantities: the position and orientation of each arm's end effector.
The state lists the current value of every quantity as a JSON object, and an action is a JSON object of the same structure that gives the change of every quantity, in centimeters and radians.
The controller moves the robot until every target is reached and the robot is at rest, or until the robot is blocked short of a target.
An episode ends when the simulator reports success, when the benchmark's step limit or the token budget of the context is reached, or when a response contains no valid action.

Because an action is a pose-level target and the controller stays outside the model, the protocol is identical for every robot: only the list of parts and the description of the robot change between LIBERO's single-arm Franka~\citep{libero}, RoboDojo's dual-arm ARX~\citep{robodojo} and the real dual-arm YAM.
Deployment on a real robot therefore requires no further training: the model acts through the same protocol, the prompt describes the real robot's frame and cameras, and the robot's own inverse kinematics and controller execute the actions.

\subsection{PEAR: a Post-training Engine for Agentic Robotics}
\label{sec:pear}

We build PEAR, an infrastructure for training RUAs, on top of verl~\citep{verl} and LMCache~\citep{lmcache}.
We train the model with reinforcement learning on robot tasks in simulation, following the recipe used for computer use~\citep{cua,operatorcard,uitars2}.
Let $\tau = (o_1, y_1, \dots, o_T, y_T)$ denote an episode, where $o_t$ is the observation of round $t$ and $y_t$ the model's response, which contains its reasoning and its action.
The policy $\pi_\theta$ generates each response token by token given the context $h_t = (p, o_1, y_1, \dots, o_t)$, where $p$ is the system prompt.
The reward $R(\tau)$ is 1 if the simulator reports success by the end of the episode and 0 otherwise, and training maximizes its expectation over the training tasks.

We use the group-normalized advantages of GRPO~\citep{deepseekmath}: we sample groups of $G$ episodes $\tau_1, \dots, \tau_G$ that share a task and an initial state and set
\begin{equation}
  \hat{A}_i = \frac{R(\tau_i) - \operatorname{mean}_j R(\tau_j)}{\operatorname{std}_j R(\tau_j) + \epsilon} ,
  \label{eq:advantage}
\end{equation}
where $\epsilon$ is a small constant.
Every generated token of an episode receives the advantage of that episode, whether it belongs to the reasoning or to an action and whichever round it is in, and the gradient estimate averages over the $N$ generated tokens of a batch:
\begin{equation}
  \begin{aligned}
    g = {} & \frac{1}{N} \sum_{i} \hat{A}_i \sum_{t, k} \Bigl[ \nabla_\theta \log \pi_\theta \bigl( y_{i,t}^{k} \mid c_{i,t}^{k} \bigr) \\
    & \quad - \mathbb{E}_{v \sim \mu ( \cdot \mid c_{i,t}^{k} )} \nabla_\theta \log \pi_\theta \bigl( v \mid c_{i,t}^{k} \bigr) \Bigr] ,
  \end{aligned}
  \label{eq:gradient}
\end{equation}
where $y_{i,t}^{k}$ is the $k$-th token of the response in round $t$ of episode $i$, $c_{i,t}^{k} = (h_{i,t}, y_{i,t}^{<k})$ is its context, and $\mu$ is the policy that generated the episode.
The first term is the REINFORCE estimator~\citep{reinforce} with the group mean as a baseline, used without PPO clipping~\citep{ppo}.
Rollouts and updates run fully asynchronously, so $\mu$ lags one or two versions behind the policy being trained.
The second term is the score-centering correction~\citep{scorecentering} for this mismatch.
We use this correction alone, without truncated importance sampling.
As in DAPO~\citep{dapo}, groups in which every episode succeeds or every episode fails are discarded and replaced because their advantages are zero.

Training is long-horizon, multi-round agentic RL, so a high prefix-cache hit rate is important for efficient rollouts.
We therefore use LMCache to offload the prefix KV cache to CPU memory and limit the number of concurrent episodes to keep the prefix cache hit rate high.
For the policy update, a whole episode forms one training sequence, which we fit in memory by combining FSDP~\citep{fsdp} with sequence parallelism~\citep{ulysses} and gradient checkpointing~\citep{gradcheckpoint}.
\Cref{app:training} gives the configuration for each type of GPU.

\subsection{Training curriculum}
\label{sec:curriculum}

With a binary reward, the advantage in \cref{eq:advantage} is nonzero only in groups that contain both successful and failed episodes, so a task on which the model never succeeds provides no learning signal.
At the start of training, almost every episode fails: the base model succeeds in about 2\% of episodes on LIBERO's Spatial, Object and Goal suites, in 0.4\% on its long-horizon suite LIBERO-Long, and almost never on most RoboDojo tasks.

We therefore train first on the tasks that already provide a signal and rely on transfer to extend it to the others.
The skills learned on the first tasks can raise the success rate on tasks that the model has not been trained on, and we add such tasks to training once the model begins to succeed on them.
In our experiments, training proceeds in three stages: LIBERO's Goal, Spatial and Object suites, then all four LIBERO suites, and then the two RoboDojo tasks that provide the strongest signal after training on LIBERO (\task{general\_pickup} and \task{align\_blocks}; \cref{sec:experiments}).
Excluding LIBERO-Long from the first stage also provides a clean test of transfer (\cref{sec:transfer}).

\section{Experiments}
\label{sec:experiments}

\textbf{Training.} We train the open-weight reasoning model Qwen3.8-27B~\citep{qwen38} with the curriculum of \cref{sec:curriculum}: 148 steps on LIBERO's Goal, Spatial and Object suites, steps 149 to 225 on all four suites, and 115 further steps on the two RoboDojo tasks.
We refer to the checkpoint at step 225 as the LIBERO-only model and label it \oursL in figures and tables.
The last checkpoint is the final model, Qwen3.8-27B-PEAR, which we deploy on the real robot and label \ours.

\textbf{Evaluation.} On LIBERO~\citep{libero}, we report the success rate over its 2,000 evaluation episodes (50 for each of its 40 tasks).
LIBERO-Plus~\citep{liberoplus} perturbs LIBERO along seven dimensions: camera, lighting, background, sensor noise, instruction wording, layout and the robot's initial state (10,030 episodes for our model).
From LIBERO-Pro~\citep{liberopro}, we use four perturbation types (2,000 episodes each for our model): a different task in the same scene, swapped object positions, a reworded instruction and restyled objects.
In RoboDojo~\citep{robodojo}, we evaluate \task{general\_pickup} and \task{align\_blocks} with 163 to 200 episodes per task and report 95\% Wilson intervals.
RoboLab-120~\citep{robolab} comprises 120 tabletop tasks at three levels of difficulty, performed by a single-arm Franka with a Robotiq gripper.
We run ten episodes per task (1,200 in total) and report the mean of two evaluation runs for the final model.
On the real dual-arm YAM, we run 30 trials for each model: two in each of five scenes with the instruction ``Use the set square to push the three blocks into a straight, aligned row, then release the set square and reset the robot arm'' and two for each of ten objects with the instruction ``Pick up the $\langle$object$\rangle$ by 10 cm.''
A trial fails if it takes more than 25 rounds or if the model reasons for more than 32k tokens in a single round.
When the model outputs an invalid action that inverse kinematics cannot solve, it retries, and the retry also counts as a round.
Unless stated otherwise, our models run at xhigh effort.

\textbf{Baselines.} Frontier vision-language models use the same protocol as our model (\cref{sec:interface}): GPT-6.1 Sol~\citep{gpt61sol} and GPT-6 Astra~\citep{gpt6astra} at the medium and max effort settings.
Effort is each model's own setting for how long it reasons.
\pizf~\citep{pi05} is a VLA supervised fine-tuned on demonstrations of every LIBERO task.
We do not evaluate \pizf ourselves and use results reported for its released checkpoint.
Its LIBERO results are from the openpi repository~\citep{openpi}, its LIBERO-Plus result is from \citet{flexpi}, and its LIBERO-Pro results are the per-suite results of \citet{liberopro} averaged over the four suites.
On RoboLab-120, we compare with four generalist robot policies from the RoboLab leaderboard~\citep{robolableaderboard}: Cosmos3-Nano-Policy~\citep{cosmos3}, \pizf, DreamZero~\citep{dreamzero} and GR00T N1.6~\citep{gr00tn16}.
All four are checkpoints trained on the real-robot dataset DROID~\citep{droid} and evaluated by the RoboLab organizers, so this \pizf is a different checkpoint from the LIBERO one above.

\subsection{Learning robot use on LIBERO}
\label{sec:libero}

RL improves the policy from a near-zero start.
On Goal, Spatial and Object, the base model succeeds in about 2\% of rollouts, so almost every rollout ends with no reward.
The training success rate on these three suites still rises to about 70\% within 148 steps (\cref{fig:libero}a).
The skills learned on these three suites also transfer well to tasks that the model has never been trained on.
At step 148, before it is trained on LIBERO-Long, the model solves 39.0\% of it, up from 0.4\% for the base model (\cref{fig:transfer}a).
Its success rate on LIBERO as a whole is 64.0\% at this point, and we return to this transfer in \cref{sec:transfer}.
Training on all four suites from step 149 then raises the success rate to 69.5\% at step 184 and to 74.4\% at step 225, each about 5 points above the last.
At 74.4\%, our model approaches GPT-6 Astra~\citep[80.6\%;][]{gpt6astra} and GPT-6.1 Sol~\citep[76.2\%;][]{gpt61sol} at max effort and exceeds both at medium effort (49.3\% and 48.1\%; \cref{fig:libero}b).

These results show that a base model with a low initial success rate can be trained to a high success rate with RL alone, from 1.6\% to 74.4\% in this case.
The capability it acquires is also general rather than specific to LIBERO's training tasks: the following sections show that the model remains robust when LIBERO's scenes and tasks are perturbed, follows instructions it was never trained on (\cref{sec:robust}), and transfers to a new benchmark, a new robot and the real world (\cref{sec:transfer,sec:sim2real}).

\subsection{Robustness and instruction following}
\label{sec:robust}

Trained only on LIBERO, our model generalizes beyond its training tasks: it remains robust when they are perturbed and follows instructions that it has never seen (\cref{fig:robust}).
\pizf~\citep{pi05} outperforms our model on LIBERO, where it is supervised fine-tuned on demonstrations of every task, but it loses this lead when a task's goal or object layout changes.
We test both policies on LIBERO-Plus~\citep{liberoplus} and LIBERO-Pro~\citep{liberopro}, which perturb the training tasks in ways that neither policy has seen.
LIBERO-Plus mainly perturbs the observation, such as the camera, the lighting, the background and sensor noise, and the two policies lose a similar amount on it: 11.2 points for \pizf and 12.2 for our model.
LIBERO-Pro also changes the goals and the object layouts.
Our model again loses about 12 points there, but \pizf loses 43.6, so its 22.5-point lead on LIBERO becomes an 8.8-point deficit (\cref{fig:robust}a).

Our model's lead on LIBERO-Pro comes from the perturbations that change what the robot has to do.
The task perturbation keeps a training scene and its initial state but gives an instruction for a different task.
\pizf falls from 96.9\% on LIBERO to 0.75\%, a drop of 96.2 points, while our model falls from 74.4\% to 57.4\%, a drop of 17.0 points (\cref{fig:robust}b).
\pizf tends to carry out the task it was trained on in that scene, while our model does what the new instruction asks.
The position perturbation tests the same ability from the other side: it keeps the instruction but swaps the objects' positions, so following the instruction now takes a different motion.
\pizf drops by 76.1 points to 20.8\%, and our model by 16.9 points to 57.5\% (\cref{fig:robust}c).

\pizf still leads on the other two perturbations, which reword the instruction or restyle the objects but leave the original task in place (\cref{fig:breakdown}b).
Thus, \pizf remains robust when the motion it was shown is still correct and fails when the motion has to change, whereas our model loses between 1 and 17 points on each perturbation but keeps following the instruction.
This difference may come from imitating demonstrations versus reasoning about the task: our model keeps the base model's ability to read an instruction and a scene, and RL teaches it to act on what it reads.

\subsection{Zero-shot transfer to unseen tasks and robots}
\label{sec:transfer}

Without training on LIBERO-Long episodes, our model reaches 39.0\% on LIBERO-Long, where the base model succeeds in 0.4\% of episodes (\cref{fig:transfer}a).
RL could hardly have learned LIBERO-Long from the base model, because a model that almost never succeeds receives almost no reward and has little to learn from.
Transfer from the other three suites changes this: when LIBERO-Long is added to training at step 149, the model's successes on it give RL a signal to train on, and its success rate on LIBERO-Long reaches 53.2\%.

The skills also transfer when both the simulator and the robot change.
RoboDojo~\citep{robodojo} is a different environment, and its tasks use a dual-arm ARX instead of LIBERO's single-arm Franka.
Trained only on LIBERO, our model succeeds in 39.9\% of RoboDojo's \task{general\_pickup} episodes and 17.0\% of its \task{align\_blocks} episodes, compared with 10.7\% and 13.0\% for the base model (\cref{fig:transfer}b).
What the model learns on LIBERO is therefore not tied to LIBERO's simulator or to its robot.

We also evaluate the final model zero-shot on RoboLab-120~\citep{robolab}, a third simulation benchmark.
It was trained with RL only on LIBERO and RoboDojo (\cref{sec:curriculum}), and RoboLab's tasks and scenes are new to it.
It succeeds in 37.3\% of RoboLab-120's 1,200 episodes, while the base model succeeds in 5.4\%.
This is on par with Cosmos3-Nano-Policy~\citep{cosmos3} at 36.8\% and above \pizf~\citep{pi05} at 28.0\%, DreamZero~\citep{dreamzero} at 25.7\% and GR00T N1.6~\citep{gr00tn16} at 7.2\%, four generalist robot policies on the RoboLab leaderboard~\citep{robolableaderboard} (\cref{fig:transfer}c).
The LIBERO-only model succeeds in 22.1\%.
\Cref{app:robolab} reports the results by task difficulty.

\subsection{Sim-to-real transfer}
\label{sec:sim2real}

Starting from the LIBERO-only model, we continue training with RL on two RoboDojo tasks, entirely in simulation.
RL works here as it does on LIBERO, on a different simulator and a dual-arm robot: in 115 steps, the training success rate on the two tasks rises from about 25\% to about 70\% (\cref{fig:sim2real}a).
Evaluated in RoboDojo, the final model succeeds in 71.2\% of \task{general\_pickup} episodes and 78.8\% of \task{align\_blocks} episodes, up from 39.9\% and 17.0\% before this stage (\cref{fig:sim2real}b).

We then evaluate the models zero-shot on a real dual-arm YAM, a robot that no model was trained on.
Without any real-robot data, the final model succeeds in 18 of 30 trials, whereas the base model succeeds in none.
The LIBERO-only model succeeds in 7 of the 30 trials, although it has seen neither these tasks nor this robot (\cref{fig:real}a and~b).
The successful trials of the final model take between 7 and 24 rounds and use between 23k and 132k tokens of context; \cref{app:real} lists every trial and the cause of each failure.

\textbf{Reasoning tasks.} The trials above repeat the two tasks on which the final model was trained in RoboDojo.
We additionally give it two reasoning tasks, which we define as tasks that have no counterpart in its training and that depend on reasoning about what to do rather than on a trained motion (\cref{fig:real}c).
For such tasks, the prompt describes the robot and the task in detail (\cref{sec:limitations}).
Given a picture of \emph{The Starry Night}, the final model draws a simple line version of the painting on a whiteboard.
To test whether a solution is alkaline, it selects phenolphthalein rather than iodine or ethanol because phenolphthalein turns pink in an alkaline solution, and the solution turns pink once the reagent is added.
Such tasks require world knowledge and reasoning ability, which stem from the foundation model.

\subsection{Retaining general capabilities}
\label{sec:general}

We compare the final model with the base model on the same questions using paired comparisons.
On all nine benchmarks~\citep{aime,gpqa,ifbench,livecodebench,realworldqa,erqa,mmmu,mmmupro}, the 95\% interval includes zero (\cref{fig:general}).
This is consistent with the finding that RL on a new task erodes a model's other skills less than supervised fine-tuning does~\citep{rlrazor}.

While reaching 75.0\% on RoboDojo's \task{general\_pickup} and \task{align\_blocks}, the final model scores 98.5 on AIME 2026~\citep{aime}, 89.7 on GPQA Diamond~\citep{gpqa} and 88.4 on LiveCodeBench~\citep{livecodebench}, and remains within two points of the base model on every vision benchmark.
The same model now reasons, writes code, answers questions about images and uses a robot.

\subsection{Reasoning effort and inference speed}
\label{sec:effort}

Although the model is trained only at xhigh effort, the effort setting in the chat template controls how long it reasons.
No RL rollout runs at a lower setting, yet at low effort the model uses 0.86k thinking tokens per round instead of 3.23k (\cref{fig:effort}a), so RL at a single effort has not fixed the length of its reasoning.
Its success rate falls with that length: on LIBERO, it is 21.9\% at low effort (the mean of two evaluation runs) and 74.4\% at xhigh.
With less than a third of the thinking, it keeps less than a third of its success (\cref{fig:effort}b).
Most of what it can do on LIBERO therefore comes from reasoning at length before each move.

Because the model reasons at length before every move, how fast it decodes determines how fast it can act.
With NVFP4 weights~\citep{nvfp4} and a DFlash2 drafter~\citep{dflash2}, it decodes 52 tokens/s on an M5 Max in Inco's Splash engine~\citep{splash}.
We optimize our megakernel with KDA~\citep{kda}.
With the megakernel, it decodes 221 tokens/s on a single RTX 5090 and 747 tokens/s on a single GB300 (\cref{fig:effort}c).
At 747 tokens/s, a round of xhigh reasoning takes about 4.3~s and a round at low effort about 1.1~s, so the effort setting also trades success for speed.
We use Inco's original DFlash2 checkpoint, with an average acceptance length of 3.2 tokens.
A drafter tuned specifically for our model may have a longer acceptance length, which would make decoding faster.
The model can therefore run locally on a single consumer GPU or a laptop.
However, NVFP4 quantization reduces our model's accuracy and lengthens its chain of thought, so all evaluations use BF16 weights.

\section{Limitations}
\label{sec:limitations}

\begin{itemize}
  \item We have not been able to scale up our training.
  Mainly because of limited GPU resources, we stopped after training on two RoboDojo tasks, and the training was stable.
  Training so far also covers only two simulators and a small set of tasks, and it starts from a 27B base model.
  We believe that scaling up, by training in more diverse simulations and on more diverse tasks and by starting from a stronger foundation model, could lead to stronger robot-use performance.
  \item We did not use supervised fine-tuning (SFT), for two reasons.
  First, we lacked high-quality in-distribution data and were concerned that SFT would harm the model's general capabilities~\citep{rlrazor,opd}.
  Second, we think that robotics data, even when available, should be added at the pretraining and mid-training stages rather than used for SFT on a model that has already been post-trained.
  We therefore train with RL alone.
  \item Our model is still slow at carrying out tasks, because it reasons at length before every move.
  We apply no length penalty during RL and place no constraint on the length of its chain of thought.
  The inference speed is high, at 747 tokens/s on a single GB300, but a round of reasoning still takes about 4.3~s on average (\cref{fig:effort}), and in a few cases the chain of thought can reach 32k tokens, so that a single move takes more than half a minute.
  Shortening the reasoning without giving up success, for example by rewarding shorter chains of thought during RL, is left to future work.
  \item Our base model, Qwen3.8-27B, is relatively small, and with its limited capabilities it often lacks knowledge about the embodiment it controls, so this information has to be provided in the prompt.
  As a result, the task success rate can depend considerably on whether the model can form a clear understanding of the embodiment and the environment from the prompt, and the model reaches its best performance when the prompt gives it enough of the information it needs.
  For harder tasks, such as those in \cref{fig:real}c, our model succeeds only after we describe the embodiment and the task to it in detail in the prompt.
  Its performance therefore depends heavily on the prompt, but this also demonstrates, to some extent, its ability to follow detailed language instructions.
  \item Our current model is not yet able to plan action chunks reliably, so it outputs the end-effector's next pose; inverse kinematics and a controller then move the arm to that pose and stop it there.
  Accuracy therefore depends partly on the controller: a poorly implemented one can move the arm abruptly at some positions and miss the target pose, and an inefficient one consumes the task's step budget and leaves fewer moves, which matters mainly in simulated evaluation.
  This design also increases the number of rounds a task needs: since each round moves the arm to a single pose, a complex motion that the model could otherwise complete in one round with one action chunk now has to be split into several rounds.
  We hope future work can improve this.
\end{itemize}

\section{Conclusion}
\label{sec:conclusion}

A general reasoning model can learn to use real robots entirely in simulation in the same way that frontier models learn to use computers.
It acts through an interface with camera frames as input and end-effector poses as output and is trained with RL on robot tasks.
After training only on LIBERO and RoboDojo in simulation, our model succeeds in 60\% of our trials on a real dual-arm robot, whereas the base model succeeds in none.
Starting from a base model that rarely succeeds at robot tasks, \rul also yields a model that follows instructions it was never trained on, remains robust when its training tasks are perturbed, transfers to a new simulator and a new robot, and retains its general capabilities.
\rul in simulation thus offers a simple recipe for real-world robot-use agents with substantial room for improvement, and our results suggest that reasoning allows such agents to generalize well beyond the tasks they were trained on.
We hope that these results encourage further work on \rul as a component of foundation-model post-training.

\section*{Acknowledgements}
We thank MIT-IBM Watson AI Lab, Amazon and MIT Science Hub, MIT AI Hardware Program, National Science Foundation, and Samsung for supporting this research.
We thank NVIDIA for donating the DGX server.
We thank Shuo Yang for his insightful discussion on model deployment.

\bibliography{refs}
\bibliographystyle{icml2026}

\newpage
\appendix
\onecolumn
\raggedbottom
\makeatletter\clubpenalty=10000 \@clubpenalty=10000 \widowpenalty=10000\makeatother
\section{Training details}
\label{app:training}

\Cref{tab:setup} lists the training setup.
Training runs on LIBERO~\citep{libero} for 225 steps and then on two RoboDojo~\citep{robodojo} tasks for 115 steps (\cref{sec:experiments}); \cref{sec:pear} describes the RL algorithm and the training infrastructure.
During training, the framework executes the JSON object of an action as a tool call to the simulator, and the result of the call is the next observation.
The controller that executes an action is specific to each robot: it drives LIBERO's operational-space controller on the Franka and follows Cartesian waypoints with inverse kinematics on RoboDojo's ARX.
The reasoning effort is xhigh in every rollout.
Earlier rounds remain in the context, and an episode ends when the simulator reports success, when the step limit or the token budget of the context is reached, or when a response contains no valid action.

\begin{table}[H]
  \caption{\textbf{Training setup.}}
  \label{tab:setup}
  \centering
  \small
  \begin{tabular}{@{}p{0.17\linewidth}p{0.80\linewidth}@{}}
    \toprule
    Setting & Value \\
    \midrule
    Base model & Qwen3.8-27B~\citep{qwen38} \\
    RL algorithm & GRPO~\citep{deepseekmath} with DAPO's group filter~\citep{dapo} and score centering~\citep{scorecentering}; a REINFORCE~\citep{reinforce} policy loss in bypass mode, without PPO clipping~\citep{ppo} \\
    Batch & 128 groups of 8 rollouts per step, counted after the group filter \\
    Optimization & AdamW~\citep{adamw} with a learning rate of $10^{-6}$ and one update per step; the loss is averaged over the generated tokens of a batch (token-mean), without a KL penalty or an entropy bonus \\
    Sampling & Temperature 1.0 over the full distribution \\
    Curriculum & LIBERO: Goal, Spatial and Object for steps 1--148; all four suites from step 149 to step 225 (the LIBERO-only model). RoboDojo: \task{general\_pickup} and \task{align\_blocks} for 115 steps from the LIBERO-only model (steps 226--340) \\
    Max sequence length & 256k tokens on LIBERO, 160k on RoboDojo \\
    Reasoning effort & xhigh throughout training \\
    Framework & verl~\citep{verl}, fully asynchronous \\
    Rollout & verl + LMCache~\citep{lmcache}. H100: tensor parallelism 4; 64 rollouts per engine at a 160k cap and 32 at a 256k cap; 95\%+ prefix-cache hits. GB300: tensor parallelism 2, without LMCache \\
    Training & H100: FSDP~\citep{fsdp} over 8 GPUs with 8-way sequence parallelism~\citep{ulysses} (\mbox{FSDP8 $\times$ SP8}), 32{,}512 tokens per GPU, FlashAttention-3~\citep{flashattention3}, activation offloading. GB300: \mbox{FSDP4 $\times$ SP4}, 69{,}632 tokens per GPU, FlashAttention-4~\citep{flashattention4}. Gradient checkpointing~\citep{gradcheckpoint} throughout; one FSDP group per node, InfiniBand between nodes \\
    Simulators & LIBERO: rendered on CPU on the rollout nodes. RoboDojo: Isaac~\citep{isaaclab} containers on separate GPU nodes \\
    \bottomrule
  \end{tabular}
\end{table}

\textbf{Infrastructure.} Our training infrastructure, PEAR (a Post-training Engine for Agentic Robotics), is built on verl~\citep{verl} and LMCache~\citep{lmcache}.
We train our models on H100, H200 and GB300 GPUs and optimize PEAR for each of them.
On H100 GPUs, we serve rollouts with tensor parallelism 4 and train the policy with FSDP~\citep{fsdp} over 8 GPUs combined with 8-way sequence parallelism~\citep{ulysses} (\mbox{FSDP8 $\times$ SP8}), so that each GPU holds 32{,}512 tokens.
On GB300 GPUs, we serve rollouts with tensor parallelism 2 and train the policy with \mbox{FSDP4 $\times$ SP4}, with 69{,}632 tokens per GPU.
Each node forms one FSDP group, and nodes communicate over InfiniBand.
To fit such long sequences in memory, we use gradient checkpointing~\citep{gradcheckpoint} throughout.
On H100 GPUs, we also offload activations and use FlashAttention-3~\citep{flashattention3}, and on GB300 GPUs we use FlashAttention-4~\citep{flashattention4}.
Each RL step uses 128 groups of 8 rollouts.
LIBERO is rendered on the CPUs of the rollout nodes, whereas RoboDojo runs in Isaac~\citep{isaaclab} containers on separate GPU nodes.

During RL on LIBERO and RoboDojo, an episode contains about 94k tokens on average (1.8k of prompt and 92k of response), 19.7 rounds and 39 to 59 images.
Episodes become shorter as training proceeds, from 119k tokens over 19.9 rounds at the start to 82k tokens over 18.5 rounds at the end.
Because each round extends the same context, most of each request is a prefix that has already been processed.
On H100 nodes, LMCache offloads the KV cache of these prefixes to CPU memory.
We set the number of concurrent rollouts so that the prefix-cache hit rate stays at about 95\% or above: each engine runs 64 rollouts concurrently when sequences are capped at 160k tokens and 32 when they are capped at 256k.
On GB300 nodes, we serve without LMCache because their host memory is too small to serve as an offload tier.
During rollout, PEAR generates about 11{,}000 tokens/s on either an H100 node with 8 GPUs or a GB300 node with 4 GPUs.
It uses speculative decoding, with the model's own multi-token prediction (MTP) layer as the drafter, which stays frozen during training.
With 3 draft tokens per step, the average acceptance length is 2.4.
We trained our model with an earlier version of PEAR, before it supported MTP, and the training took about 50{,}000 H100-equivalent GPU-hours.
The current PEAR would bring this down to an estimated 30{,}000.

RL runs fully asynchronously in verl, so the policy that generates a rollout lags one or two versions behind the policy being trained.
We correct for this mismatch with score centering~\citep{scorecentering}, which subtracts from each token's score function its expected value under the rollout policy's next-token distribution.
We use score centering alone, without truncated importance sampling.

\Needspace{23.5\baselineskip}
\section{LIBERO by suite and effort}
\label{app:libero}

\begin{figure}[H]
  \centering
  \includegraphics[width=\textwidth]{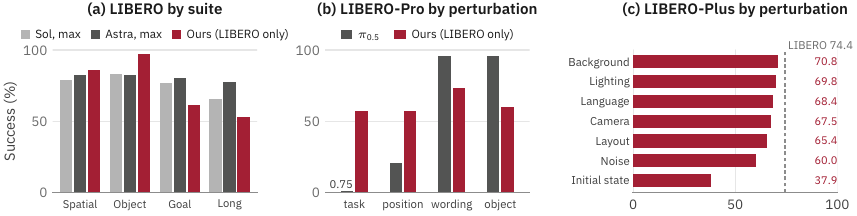}
  \caption{\textbf{LIBERO by suite, and LIBERO-Pro and LIBERO-Plus by perturbation.}
  (a) Success rate (\%) on LIBERO by suite, against the frontier VLMs at max effort (\cref{tab:libero-suites}).
  (b) Success rate (\%) on LIBERO-Pro by perturbation type: an unseen task in a training scene, swapped object positions, a reworded instruction and restyled objects (\cref{tab:libero-pro}).
  (c) The LIBERO-only model on LIBERO-Plus by perturbation dimension (10,030 episodes); the dashed line marks its success rate on LIBERO.}
  \label{fig:breakdown}
\end{figure}

\Cref{tab:libero-suites} breaks down the LIBERO results of \cref{fig:libero}b and \cref{fig:effort}b by suite, and \cref{fig:breakdown}a shows them at max effort.
Among the vision-language models, the LIBERO-only model has the highest Spatial and Object scores; on Goal and LIBERO-Long, its scores fall between those of the frontier models at medium and at max effort, and 19\% and 33\% of its episodes on these two suites end when the context reaches the 256k-token budget.
\begin{table}[H]
  \caption{\textbf{LIBERO success rate (\%) by suite.}
  The low-effort row is the mean of two evaluation runs.
  The \pizf row is from the openpi repository~\citep{openpi}.
  Bold: best VLM in each column, among the max and xhigh rows and among the medium rows.}
  \label{tab:libero-suites}
  \centering
  \small
  \begin{tabular}{llccccc}
    \toprule
    Model & Effort & Spatial & Object & Goal & Long & Average \\
    \midrule
    \multicolumn{7}{l}{\textit{Vision-language models}} \\
    Qwen3.8-27B & xhigh & 1.2 & 1.8 & 2.8 & 0.4 & 1.6 \\
    GPT-6.1 Sol~\citep{gpt61sol} & max & 78.8 & 83.4 & 76.8 & 65.8 & 76.2 \\
    GPT-6 Astra~\citep{gpt6astra} & max & 82.6 & 82.2 & \textbf{80.2} & \textbf{77.4} & \textbf{80.6} \\
    GPT-6.1 Sol & medium & 22.8 & \textbf{69.2} & 58.2 & 42.2 & 48.1 \\
    GPT-6 Astra & medium & \textbf{26.0} & 67.0 & \textbf{59.4} & \textbf{44.6} & \textbf{49.3} \\
    \midrule
    \multicolumn{7}{l}{\textit{Vision-language-action model, supervised fine-tuned on LIBERO}} \\
    \pizf & -- & 98.8 & 98.2 & 98.0 & 92.4 & 96.9 \\
    \midrule
    \multicolumn{7}{l}{\textit{This work}} \\
    \oursL & xhigh & \textbf{85.8} & \textbf{97.0} & 61.4 & 53.2 & 74.4 \\
    \oursL & low & 30.4 & 35.9 & 18.9 & 2.5 & 21.9 \\
    \bottomrule
  \end{tabular}
\end{table}

\Needspace{25.5\baselineskip}
\section{Robustness by perturbation and by suite}
\label{app:robust}

\Cref{tab:libero-pro} reports LIBERO-Pro~\citep{liberopro} results by perturbation type, on which \cref{fig:robust}b and~c and \cref{fig:breakdown}b are based.
\Cref{tab:suite-robust} reports the success rate of the LIBERO-only model by suite on LIBERO and the two perturbed sets.
On LIBERO-Plus~\citep{liberoplus} (\cref{fig:breakdown}c), the LIBERO-only model is most sensitive to the robot's initial state, which costs it 36.5 points.

\begin{table}[H]
  \caption{\textbf{LIBERO-Pro success rate (\%) by perturbation type}, both policies trained on LIBERO; 2,000 episodes per type for our model.
  The \pizf row averages the per-suite results of \citet[Tables 2--5 of the first arXiv version]{liberopro} over the four suites.
  Bold: best in each column.
  Task: the same scene with a different task.
  Position: the objects' positions swapped.
  Semantic: the instruction reworded.
  Object: the objects restyled.
  Mean: over the four types.}
  \label{tab:libero-pro}
  \centering
  \small
  \begin{tabular}{lccccc}
    \toprule
    Model & Task & Position & Semantic & Object & Mean \\
    \midrule
    \pizf, supervised fine-tuned on LIBERO & 0.75 & 20.8 & \textbf{95.8} & \textbf{96.0} & 53.3 \\
    \oursL, xhigh effort & \textbf{57.4} & \textbf{57.5} & 73.3 & 60.2 & \textbf{62.1} \\
    \bottomrule
  \end{tabular}
\end{table}

\begin{table}[H]
  \caption{\textbf{The LIBERO-only model by suite, on LIBERO and on the two perturbed sets} (success rate, \%); perturbations reduce every suite's success rate by 7--15 points.
  LIBERO-Pro's numbers include its unperturbed tasks (64.5 on average; 62.1 over the four perturbations alone).
  LIBERO-Pro results by perturbation type are in \cref{tab:libero-pro}, and LIBERO-Plus results by perturbation dimension in \cref{fig:breakdown}c.}
  \label{tab:suite-robust}
  \centering
  \small
  \begin{tabular}{lccc}
    \toprule
    Suite & LIBERO & LIBERO-Plus & LIBERO-Pro \\
    \midrule
    Spatial & 85.8 & 75.7 & 78.9 \\
    Object & 97.0 & 82.7 & 86.4 \\
    Goal & 61.4 & 46.8 & 48.7 \\
    Long & 53.2 & 44.5 & 43.8 \\
    \bottomrule
  \end{tabular}
\end{table}

\Needspace{24.5\baselineskip}
\section{RoboLab-120 by task difficulty}
\label{app:robolab}

\Cref{tab:robolab} breaks down the RoboLab-120~\citep{robolab} results of \cref{sec:transfer} by task difficulty and adds RoboLab's Score, which gives partial credit for the subtasks that a policy completes.
Among the policies in \cref{tab:robolab}, the final model achieves the highest success rate on simple and moderate tasks.
On complex tasks, Cosmos3-Nano-Policy~\citep{cosmos3} leads with 25.3\%, whereas the final model reaches 13.2\%.
The LIBERO-only model improves over the base model mainly on simple and moderate tasks (from 5.2\% to 28.6\% and from 7.7\% to 19.5\%).

\begin{table}[H]
  \caption{\textbf{RoboLab-120 by task difficulty.}
  Success rate (\%) on all 120 tasks and by difficulty (64 simple, 39 moderate and 17 complex tasks, ten episodes each), and Score, RoboLab's graded measure of progress through a task's subtasks, where a success scores 100.
  The final model is trained with RL on LIBERO and RoboDojo, and its values are the mean of two evaluation runs (2,400 episodes).
  Other policies: the RoboLab leaderboard~\citep[data of 29 October 2026]{robolableaderboard}.
  Bold: best in each column.}
  \label{tab:robolab}
  \centering
  \small
  \begin{tabular}{lccccc}
    \toprule
    Model & All & Simple & Moderate & Complex & Score \\
    \midrule
    \multicolumn{6}{l}{\textit{Our base model}} \\
    Qwen3.8-27B & 5.4 & 5.2 & 7.7 & 1.2 & 12.1 \\
    \midrule
    \multicolumn{6}{l}{\textit{Generalist robot policies}} \\
    Cosmos3-Nano-Policy~\citep{cosmos3} & 36.8 & 40.6 & 35.4 & \textbf{25.3} & \textbf{51.9} \\
    \pizf~\citep{pi05} & 28.0 & 29.7 & 31.5 & 13.5 & 43.4 \\
    DreamZero~\citep{dreamzero} & 25.7 & 26.1 & 30.0 & 14.1 & 39.8 \\
    GR00T N1.6~\citep{gr00tn16} & 7.2 & 8.8 & 7.9 & 0.0 & 17.1 \\
    \midrule
    \multicolumn{6}{l}{\textit{This work}} \\
    \oursL & 22.1 & 28.6 & 19.5 & 3.5 & 35.0 \\
    \ours & \textbf{37.3} & \textbf{44.6} & \textbf{35.8} & 13.2 & 51.3 \\
    \bottomrule
  \end{tabular}
\end{table}

\Needspace{21\baselineskip}
\section{The full RoboDojo benchmark}
\label{app:robodojo}

\begin{figure}[H]
  \centering
  \includegraphics[scale=1.2]{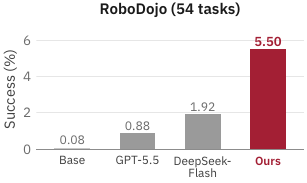}
  \caption{\textbf{Average success rate on the full RoboDojo benchmark.}
  All four models are scored with RoboDojo's official rule, the mean success rate of its five capability dimensions.
  The base model and the final model are evaluated with 10 episodes per task setting.
  GPT-5.5 and DeepSeek-Flash are entries of the leaderboard~\citep{robodojoagent} from the RoboDojo team's evaluation of language models as policies~\citep{astrarobodojo}, with 50 and 10 episodes per task.}
  \label{fig:robodojo-full}
\end{figure}

\Cref{fig:robodojo-full} shows the average success rate on the full RoboDojo simulation benchmark~\citep{robodojo}, of which \cref{sec:sim2real} reports two tasks.
The benchmark has 42 tasks in five capability dimensions: Generalization, Memory, Long-Horizon, Precision and Open.
Its 12 Generalization tasks are evaluated in a standard and a randomized setting, which gives 54 task settings, and we run 10 episodes of each, 540 episodes in total.
We follow RoboDojo's official scoring rule: the success rate measures binary task completion, the success rate of a dimension is the mean over its tasks, and the overall success rate is the mean of the five dimensions rather than of all tasks.
The leaderboard entries of GPT-5.5 and DeepSeek-Flash are scored with the same rule, so their results can be compared with ours.

The base model succeeds in 1 of the 540 episodes, on \task{stack\_blocks\_random}.
The final model succeeds in 21 episodes: 8 of 10 on \task{general\_pickup} and 7 of 10 on \task{align\_blocks}, the two tasks used in training, and 3 of 10 on each of \task{match\_and\_pick\_from\_conveyor} and \task{solve\_equation}.

A large part of the failures is due to the length of RoboDojo's tasks.
The context window of our model is 256k tokens, and 370 of the final model's 540 episodes (68.5\%) end in failure because this window is exhausted.
We believe that a stronger foundation model with a longer context window, trained with \rul, or a harness that manages the context could reach higher success rates.

\Needspace{19\baselineskip}
\section{General benchmarks}
\label{app:general}

\Cref{tab:general-items} lists the numbers behind \cref{fig:general}, together with the category of each benchmark and its number of questions and samples per question.

\begin{table}[H]
  \caption{\textbf{General benchmarks.}
  The final model and the base model on the same questions, with paired 95\% intervals and the number of questions (and samples per question).
  The intervals are from a paired bootstrap over questions with 2,000 resamples.
  evalscope~\citep{evalscope} 1.13.0, vLLM~\citep{vllm} in BF16, thinking-mode sampling (temperature 1.0, top-$p$ 0.95, top-$k$ 20).
  Changes and means are computed before rounding.
  Nine-benchmark means: 81.64 for the base model and 81.49 for the final model.
  For reference, the LIBERO-only model averages 81.31.}
  \label{tab:general-items}
  \centering
  \small
  \begin{tabular}{llcccc c}
    \toprule
    Benchmark & Category & Questions & Base & \ours & Change & 95\% CI \\
    \midrule
    AIME 2026 & math & 30$\times$16 & 98.3 & 98.5 & $+0.2$ & [$-1.7$, $+2.9$] \\
    GPQA Diamond & science & 198$\times$8 & 89.6 & 89.7 & $+0.1$ & [$-1.0$, $+1.2$] \\
    IFBench & instruction following & 300, prompt strict & 73.0 & 72.7 & $-0.3$ & [$-4.0$, $+3.3$] \\
    LiveCodeBench v6 & coding & 131$\times$4 & 88.0 & 88.4 & $+0.4$ & [$-2.1$, $+2.9$] \\
    RealWorldQA & visual QA & 765 & 84.1 & 83.7 & $-0.4$ & [$-2.4$, $+1.6$] \\
    ERQA & embodied reasoning & 400 & 68.5 & 66.5 & $-2.0$ & [$-6.2$, $+2.2$] \\
    MMMU (val) & multimodal & 900 & 80.7 & 81.0 & $+0.3$ & [$-1.3$, $+2.0$] \\
    MMMU-Pro standard & multimodal & 1730, 10 options & 77.3 & 77.6 & $+0.3$ & [$-1.1$, $+1.7$] \\
    MMMU-Pro vision & multimodal & 1730 & 75.4 & 75.4 & $+0.1$ & [$-1.4$, $+1.5$] \\
    \bottomrule
  \end{tabular}
\end{table}

\Needspace{54\baselineskip}
\section{Real-robot trials}
\label{app:real}

\Cref{tab:real} lists every real-robot trial and the cause of each failure.
All three models use the same protocol on the same robot at xhigh effort, and none of them was trained on this robot or on any real-robot data.
The final model was trained on the two corresponding tasks in RoboDojo with a different robot, whereas the LIBERO-only model and the base model were not.
The final model succeeds in 18 of 30 trials, the LIBERO-only model in 7 and the base model in none.
The successful trials of the final model take between 7 and 24 rounds, and its failed trials end after between 6 and 25 rounds.

\textbf{Failure causes.} A trial fails when it exceeds 25 rounds or 32k reasoning tokens in a single round (\cref{sec:experiments}), or when the model does not complete the task.
The base model runs out of rounds in 25 of its 30 trials, knocks the object over in 4 and finishes with the blocks not aligned in 1.
The 23 failures of the LIBERO-only model are more varied: it runs out of rounds in 10 trials, knocks the object over in 4, reasons for more than 32k tokens in a single round in 4, stops automatically in an unsafe pose in 2, does not fully lift the object in 2 and picks up the wrong object in 1.
The final model shows none of these last four kinds of failure: in its 12 failed trials, it runs out of rounds in 6, finishes with the blocks not aligned in 3 and knocks the object over in 2, and the tennis ball rolls away in 1.
No model picks up the air pump, which is knocked over in all 6 of its trials, or the brush, with which 5 of the 6 trials run out of rounds.

\begin{table}[H]
  \caption{\textbf{Every real-robot trial}, all models at xhigh effort on a dual-arm YAM.
  Align blocks: two trials in each of five scenes, which differ in the initial layout of the blocks and the set square, all with the instruction ``Use the set square to push the three blocks into a straight, aligned row, then release the set square and reset the robot arm.''
  General pickup: two trials per object, ``Pick up the $\langle$object$\rangle$ by 10 cm.''
  \textbf{S}: success within the budget of 25 rounds; F: failure, followed by its cause.
  \emph{Rounds}: ran out of rounds; \emph{32k tokens}: reasoned for more than 32k tokens in a single round; \emph{toppled}: knocked the object over; \emph{unaligned}: finished with the blocks not aligned; \emph{not lifted}: did not fully lift the object; \emph{wrong object}: picked up the wrong object; \emph{rolled}: the ball rolled away; \emph{unsafe}: stopped automatically in an unsafe pose.
  Rounds: the number of rounds, at most 25.
  Context: tokens used by the last of these rounds, out of 260,096.}
  \label{tab:real}
  \centering
  \small
  \setlength{\tabcolsep}{3.6pt}
  \begin{tabular}{lc lrr lrr lrr}
    \toprule
    & & \multicolumn{3}{c}{Qwen3.8-27B (base)} & \multicolumn{3}{c}{\oursL} & \multicolumn{3}{c}{\ours} \\
    \cmidrule(lr){3-5}\cmidrule(lr){6-8}\cmidrule(lr){9-11}
    Scene or object & Trial & Result & Rounds & Context & Result & Rounds & Context & Result & Rounds & Context \\
    \midrule
    \multicolumn{11}{l}{\textit{Align blocks}} \\
    Scene 1 & 1 & F: unaligned & 12 & 40.0k & F: unsafe & 20 & 78.2k & \textbf{S} & 10 & 54.8k \\
     & 2 & F: rounds & 25 & 77.7k & \textbf{S} & 24 & 57.8k & F: unaligned & 9 & 34.5k \\
    Scene 2 & 1 & F: rounds & 25 & 103.4k & F: 32k tokens & 10 & 99.4k & \textbf{S} & 14 & 47.1k \\
     & 2 & F: rounds & 25 & 60.6k & F: rounds & 25 & 106.0k & \textbf{S} & 8 & 26.1k \\
    Scene 3 & 1 & F: rounds & 25 & 69.6k & F: rounds & 25 & 131.8k & \textbf{S} & 18 & 80.0k \\
     & 2 & F: rounds & 25 & 54.6k & F: rounds & 25 & 72.0k & F: rounds & 25 & 119.6k \\
    Scene 4 & 1 & F: rounds & 25 & 72.1k & F: rounds & 25 & 121.0k & F: unaligned & 8 & 44.5k \\
     & 2 & F: rounds & 25 & 176.4k & F: rounds & 25 & 117.2k & \textbf{S} & 20 & 72.0k \\
    Scene 5 & 1 & F: rounds & 25 & 66.2k & F: rounds & 25 & 136.8k & \textbf{S} & 13 & 40.1k \\
     & 2 & F: rounds & 25 & 141.8k & F: rounds & 25 & 98.5k & F: unaligned & 11 & 40.7k \\
    \addlinespace[3pt]
    \multicolumn{11}{l}{\textit{General pickup}} \\
    Air pump & 1 & F: toppled & 5 & 17.6k & F: toppled & 6 & 22.0k & F: toppled & 7 & 28.9k \\
     & 2 & F: toppled & 3 & 7.4k & F: toppled & 13 & 33.4k & F: toppled & 6 & 20.6k \\
    Brush & 1 & F: rounds & 25 & 66.0k & F: unsafe & 19 & 86.4k & F: rounds & 25 & 107.5k \\
     & 2 & F: rounds & 25 & 71.8k & F: rounds & 25 & 126.2k & F: rounds & 25 & 92.1k \\
    Dishwashing liquid & 1 & F: rounds & 25 & 40.0k & F: 32k tokens & 15 & 79.7k & F: rounds & 25 & 108.3k \\
     & 2 & F: rounds & 25 & 74.8k & F: 32k tokens & 17 & 131.4k & \textbf{S} & 13 & 33.8k \\
    Stapler & 1 & F: rounds & 25 & 85.0k & F: wrong object & 10 & 59.7k & \textbf{S} & 13 & 52.5k \\
     & 2 & F: rounds & 25 & 103.1k & F: rounds & 25 & 93.4k & F: rounds & 25 & 138.8k \\
    Remote control & 1 & F: rounds & 25 & 100.0k & F: rounds & 25 & 157.9k & \textbf{S} & 14 & 48.9k \\
     & 2 & F: rounds & 25 & 87.7k & \textbf{S} & 21 & 76.5k & \textbf{S} & 7 & 23.5k \\
    Rope & 1 & F: rounds & 25 & 75.4k & \textbf{S} & 22 & 135.7k & \textbf{S} & 9 & 47.6k \\
     & 2 & F: rounds & 25 & 38.4k & \textbf{S} & 7 & 15.2k & \textbf{S} & 9 & 25.6k \\
    Screw & 1 & F: rounds & 25 & 46.6k & \textbf{S} & 13 & 31.8k & \textbf{S} & 18 & 82.3k \\
     & 2 & F: rounds & 25 & 77.6k & \textbf{S} & 11 & 33.4k & \textbf{S} & 24 & 132.0k \\
    Spray can & 1 & F: toppled & 16 & 51.9k & F: toppled & 6 & 21.1k & F: rounds & 25 & 129.4k \\
     & 2 & F: toppled & 15 & 31.3k & F: toppled & 8 & 21.5k & \textbf{S} & 8 & 36.8k \\
    Tennis ball & 1 & F: rounds & 25 & 60.8k & F: 32k tokens & 15 & 141.1k & \textbf{S} & 22 & 91.5k \\
     & 2 & F: rounds & 25 & 90.3k & \textbf{S} & 19 & 69.1k & F: rolled & 17 & 81.9k \\
    Turtle plush toy & 1 & F: rounds & 25 & 54.2k & F: not lifted & 11 & 33.1k & \textbf{S} & 15 & 28.2k \\
     & 2 & F: rounds & 25 & 67.3k & F: not lifted & 10 & 27.3k & \textbf{S} & 10 & 23.6k \\
    \midrule
    Successes & & 0/30 & & & 7/30 & & & \textbf{18/30} & & \\
    \bottomrule
  \end{tabular}
\end{table}

\Needspace{42.5\baselineskip}
\section{Prompts}
\label{app:prompts}

The system prompt consists of a protocol that is the same for every environment, an ENVIRONMENT section that is specific to the benchmark, and the task.
The template is given below, where \texttt{\{environment\}} and \texttt{\{instruction\}} stand for the benchmark's ENVIRONMENT section and the task instruction and all other text is given to the model as shown.

\begin{lstlisting}[style=prompt]
You are controlling a robot. Each turn you see the robot's current state and its camera images, and you respond with the next action.

## OBSERVATION

The robot is made of named parts, and each part has named quantities you can set. Every turn begins with the current value of every such quantity, grouped by part, followed by the camera images:

{"<part>": {"<quantity>": <value>, ...}, ...}

Lengths are in centimeters and angles in radians; any other quantity is a plain number. Nothing beyond these names is given about what each quantity corresponds to.

## ACTION

The action changes the listed quantities by these amounts:

{"<part>": {"<quantity>": <change>, ...}, ...}

Write the change of every quantity; 0 leaves a quantity as it is.

## EXECUTION

The action ends once every target is reached and the robot is at rest. If something keeps a target from being reached, the robot stops there and the action ends. Either way, you then receive the next observation.

## RESPONSE FORMAT

End your response with exactly one fenced json block containing the action, and nothing after it:

```json
<action>
```

## ENVIRONMENT

{environment}

## TASK

{instruction}
\end{lstlisting}

\Needspace{8\baselineskip}
LIBERO's ENVIRONMENT section describes its two cameras.

\begin{lstlisting}[style=prompt]
- agentview: a fixed camera; the robot's base is at the top of its image, +x points toward the bottom of the image and +y toward its left.
- wrist: a camera on the gripper; with the gripper pointing straight down at yaw 0, +x points toward the top of its image and +y toward its right.
- the episode ends by itself as soon as the task is done: while observations keep coming, the task is not done yet.
\end{lstlisting}

\Needspace{33.5\baselineskip}
RoboDojo's ENVIRONMENT section describes the frame of the two arms, the orientation of the grippers and the three cameras.
On the real YAM, the prompt follows the same template with this section.

\begin{lstlisting}[style=prompt]
Robot and frame:
- left_arm and right_arm share one frame. The origin is midway between the two arms' bases, at the height they are mounted at; +x points forward, +y to the robot's left, +z up.
- left_arm works on the +y side, right_arm on the -y side.

Gripper orientation:
- Both grippers start pointing forward (roll -1.57, pitch 0, yaw 1.57).
- To grasp from above, first turn the gripper to point down: from the start, a roll change of +1.57 points it straight down.
- Pointing down, yaw turns the gripper about the vertical: at yaw 1.57 its fingers sit on its left and right and pinch along y; at yaw 0 they sit in front and behind and pinch along x.

Cameras:
- head: fixed above the robot, looking forward and down. Up in the image is +x (farther from the robot), left is +y.
- left_wrist, right_wrist: one on each gripper, looking along its fingers, which show at the left and right edges of the image. With the gripper pointing straight down at yaw 1.57, up in the image is +x and left is +y.

Ending:
- The episode ends by itself as soon as the task is done: while observations keep coming, the task is not done yet.
- Unless the task is to end up holding something, it is done only once both arms are back near the values they started with.

Answer format:
- Every action lists all four parts with every quantity; write 0 for whatever should not change. For example, this action only lowers the right gripper by 5 and closes it:
```json
{"left_arm": {"x": 0, "y": 0, "z": 0, "roll": 0, "pitch": 0, "yaw": 0}, "right_arm": {"x": 0, "y": 0, "z": -5, "roll": 0, "pitch": 0, "yaw": 0}, "left_gripper": {"width": 0}, "right_gripper": {"width": -1}}
```
\end{lstlisting}

\Needspace{22\baselineskip}
RoboLab's ENVIRONMENT section describes the frame of the arm, the gripper and the two cameras.

\begin{lstlisting}[style=prompt]
Robot and frame:
- The arm's frame has its origin at the robot's base, at the height of the table top: +x points forward, +y to the robot's left, +z up.
- The arm's position is the point between the gripper's finger pads, about 2 above the fingertips.

Gripper:
- Pointing straight down is roll 3.14 (or -3.14) and pitch 0; pointing down, yaw turns the gripper about the vertical, and at yaw 0 its fingers pinch along y. Yaws 3.14 apart pinch alike, and the wrist turns a little less than half a turn either way from yaw 0: of two such yaws, turn to the one nearer 0.
- Open, the fingers are about 8 cm apart. A close lasts until an open: when it left the fingers wide apart, closing again changes nothing. Closing also draws the gripper back a little along its axis, so the fingers close where they were.

Cameras:
- over_shoulder: fixed on the robot's left, looking forward and down across the table; +x points toward the upper left of the image and +y toward its left.
- wrist: on the gripper, behind the fingers and to their right, looking down past them; with the gripper pointing down at yaw 0, +x points toward the bottom of the image and +y toward its right, and the spot straight below the fingers lies right of the image centre, and above it while the gripper is high.

Ending:
- The episode ends by itself as soon as the task is done.
- The task has a time limit, and every action spends some of it.
\end{lstlisting}

\end{document}